\documentclass[11pt]{article}

\usepackage{acl}

\usepackage{times}
\usepackage{latexsym}

\usepackage[T1]{fontenc}

\usepackage[utf8]{inputenc}

\usepackage{microtype}

\usepackage{graphicx}
\usepackage{algorithm}
\usepackage{algpseudocode}
\usepackage{multirow}
\usepackage{amsmath}
\usepackage{mathtools}
\usepackage{xurl}
\usepackage[table]{xcolor}
\title{KV-Repurpose: A Testbed for Evaluating Non-Prefix KV-Cache Reuse}

\author{Taiga Ikeda \and Daisuke Miyashita \and Jun Deguchi \\
        Kioxia Corporation, Japan}

\begin{document}

\maketitle
\begin{abstract}
KV-cache reuse in large language model serving is commonly limited to exact prompt prefixes, preventing reuse when the same retrieved document appears after different system prompts, conversation histories, or dynamic metadata. We present KV-Repurpose, an open-source, vLLM-based system for developing and evaluating segment-wise KV-cache reuse beyond exact prefixes. The system combines cache-aware segment scheduling, prompt construction, capacity management, workload generation, and runtime monitoring. Experiments on long-context benchmarks show that naive segment reuse largely preserves output quality for single-document question answering and summarization, while exposing clear limitations in cross-document reasoning. Under a Zipfian RAG-style workload in which 20\% of requests share a fixed prefix, KV-Repurpose achieves a substantially higher token-level cache hit rate than prefix caching at the same 30 GB cache capacity, 63.9\% versus 3.3\%, resulting in lower TTFT. Even with 1 TB of cache capacity, prefix caching reaches only a 17.9\% hit rate in this workload. Source code is available under the Apache-2.0 license at \url{https://github.com/taiga333/eval-kv-repurpose}.
\end{abstract}


\section{Introduction}
Large language models (LLMs) are now widely deployed in applications such as conversational assistants, document summarization, retrieval-augmented generation (RAG), and code generation.
As recent models have become capable of processing increasingly long inputs, demand for long-context inference in practical deployments has grown rapidly.
For many LLM serving workloads, however, processing long prompts remains computationally expensive and can substantially increase time to first token (TTFT)~\citep{promptcache}.

To avoid repeated prefill computation across requests, modern LLM serving engines employ prefix caching~\citep{vllm,sglang}.
Prefix caching stores the key-value (KV) states computed for previously processed prefixes and reuses them when subsequent requests share the same prefix.
It is particularly effective in workloads with recurring prefixes, such as multi-turn conversations~\citep{MultiTurn} and in-context learning~\citep{icl}.

By contrast, in RAG workloads~\citep{rag,rag_survey}, the composition and ordering of retrieved documents often vary across requests.
As a result, requests may contain substantial overlapping content while failing to share an exact prefix.
Widely used serving engines~\citep{lmcache}, however, primarily optimize exact-prefix reuse and do not yet support the reuse of the corresponding KV states in such cases \footnote{\url{https://github.com/LMCache/LMCache/issues/3238}\label{issue}}, potentially leading to redundant computation, increased TTFT, and reduced inference throughput.
It can also increase pressure on KV-cache capacity across the memory and storage hierarchy, including dedicated context-memory systems~\cite{nvidia_cmx}.

\begin{figure}[t]
  \includegraphics[width=\columnwidth]{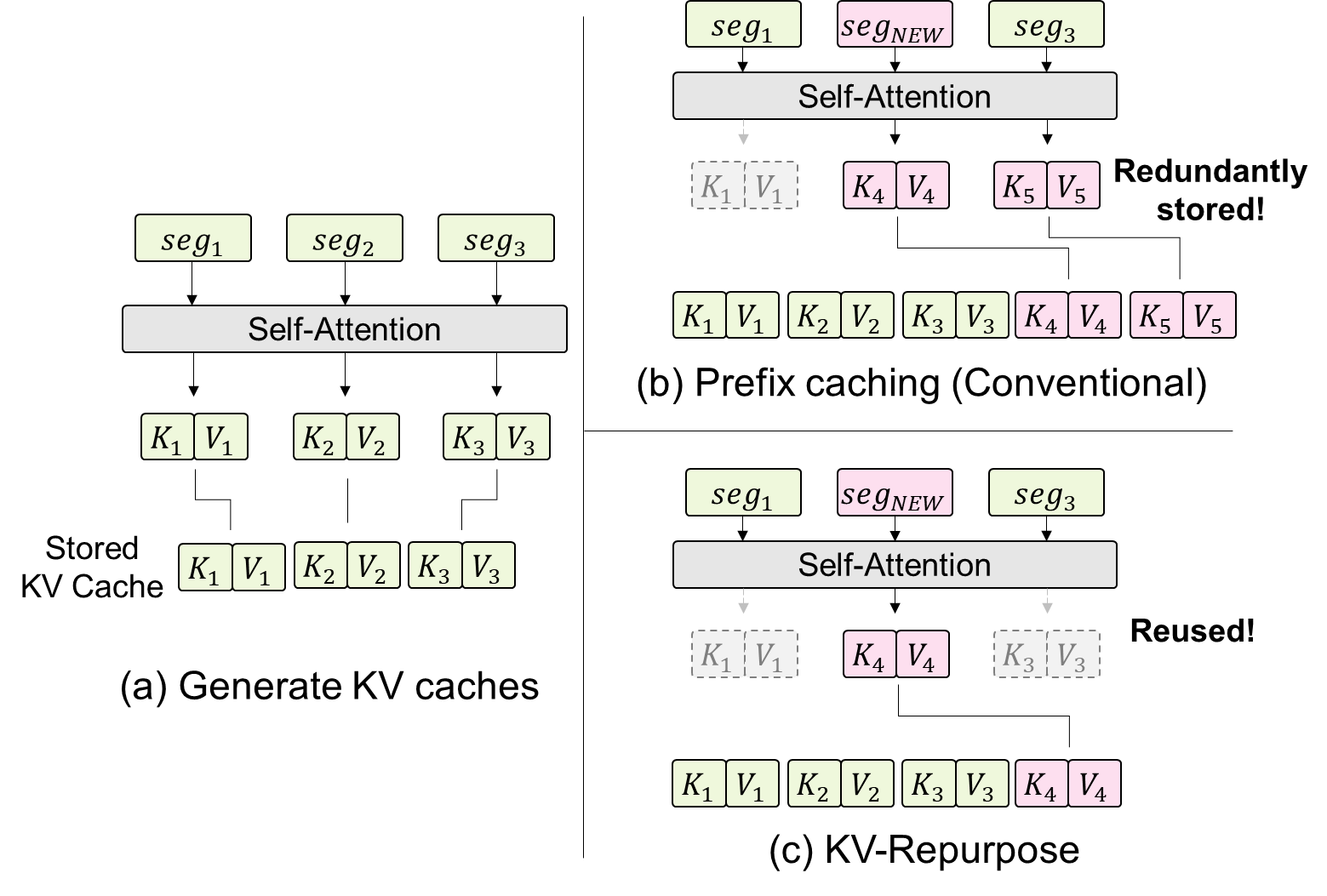}
  \caption{Motivation.}
  \label{fig:motivation}
\end{figure}

Prior work has explored techniques for preserving or recovering model quality when cached KV states are reused beyond exact-prefix matches~\citep{cachecraft}.
Building on this line of research, we revisit such reuse from a systems perspective and broadly refer to it as \textit{KV-Repurpose} in this paper (Figure~\ref{fig:motivation}).
Our goal is neither to claim to be the first to introduce the concept of repurposing nor to propose a state-of-the-art method for mitigating the accuracy degradation caused by repurposing.
Rather, we seek to reassess the practical value of KV-Repurpose in large-scale, long-context inference.

However, simple and reproducible baselines for integrating non-prefix KV-cache reuse into modern serving engines remain limited, making it difficult to assess its practical value and compare more sophisticated techniques on a common basis. 
We therefore develop a testbed for naive KV-Repurpose that enables systematic evaluation of both serving performance and output quality in long-context inference settings.
Using this system, we demonstrate the potential benefits of KV-Repurpose through representative synthetic workloads motivated by RAG deployments.
Our contributions are as follows:

\begin{itemize}
  \item We integrate segment-wise, non-prefix KV-cache reuse into vLLM as a research baseline.
  \item We develop an end-to-end evaluation system covering cache generation, prompt construction, eviction, and runtime monitoring.
  \item We empirically characterize the conditions under which naive reuse preserves model quality and those under which it fails.
  \item We evaluate the system on a model with more than 100 billion parameters and demonstrate substantial improvements in cache-capacity efficiency and latency under long-context, Zipfian RAG-style workloads.
\end{itemize}

\section{System Implementation}
\begin{figure}[t]
  \centering
  \includegraphics[width=\columnwidth]{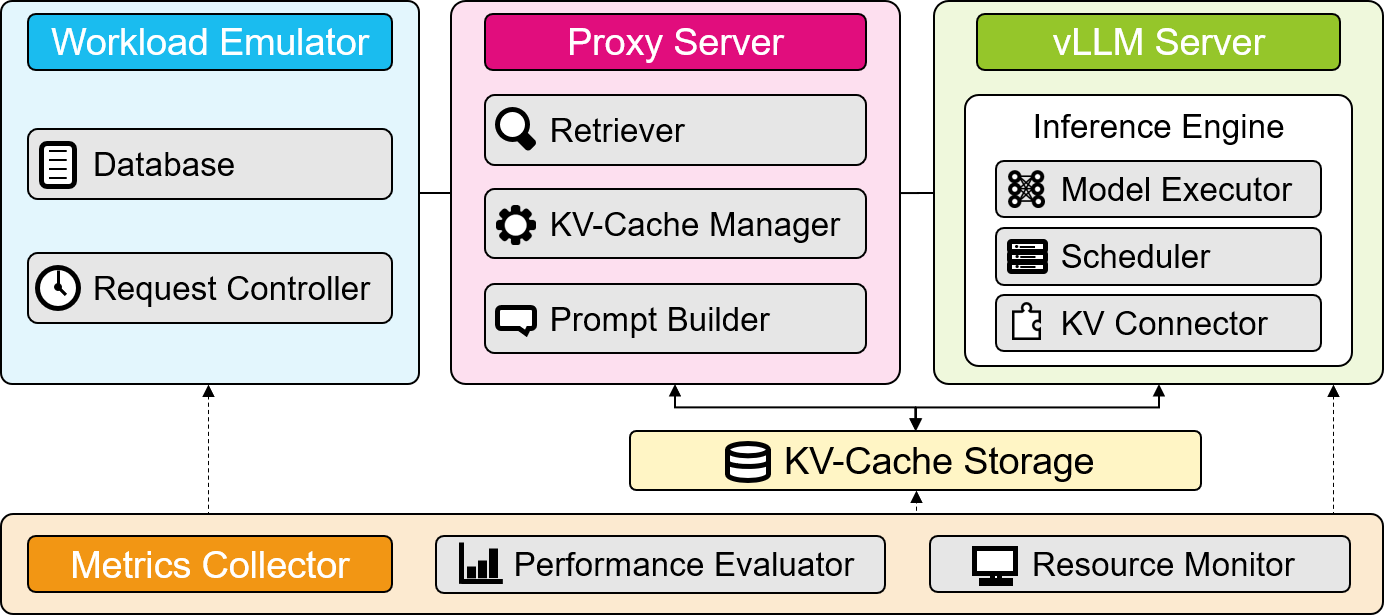}
  \caption{System overview.}
  \label{fig:overview}
\end{figure}

This section describes the implementation of the proposed system, illustrated in Figure~\ref{fig:overview}.
The system comprises two main components: an extended vLLM server for KV-Repurpose and a proxy server for prompt building and KV-cache management.
The former provides the main inference service, while the latter supports request preprocessing and cache-related operations for performance evaluation.

The system also includes modules for evaluation and monitoring.
It supports task-level evaluation on long-context comprehension tasks, workload emulation with GuideLLM~\citep{guidellm} to measure metrics such as TTFT and throughput, and monitoring with Prometheus/Grafana~\citep{vllm_prometheus_grafana} to track cache hit rates and storage utilization.
The following subsections detail the design, implementation, and usage of the extended vLLM server and the proxy server.

\subsection{Extensions to the vLLM Server}
LLM inference engines such as vLLM, including KV-cache extensions like LMCache, are mainly optimized for prefix caching under exact prefix match.
To address this limitation, we extend vLLM with segment-wise cache-aware scheduling and a lightweight implementation of position-adjusted KV reuse based on an approach commonly adopted in prior work~\citep{minipic,mepic}.

First, we introduce segment-wise cache-aware scheduling. 
The input is divided by special delimiter tokens and processed incrementally by segment.
For each segment, the server independently checks KV-cache availability: if available, it skips prefill and reuses the cached states; otherwise, it performs prefill and stores the resulting KV states. 
Unlike most existing systems, reuse is determined independently for each segment, so a miss in one segment does not prevent reuse in later segments.

Second, we store each segment's keys in an unrotated form and apply RoPE according to their current positions upon reuse. 
This straightforward adjustment only realigns the cached keys with their new positions and does not recompute attention with the preceding prefix.

\paragraph{Usage}
The extended server can be launched in the same way as a standard vLLM server using the \texttt{vllm serve} command, with an explicit KV-transfer configuration such as \texttt{\{"kv\_connector":"RepurpConnector"\}}.
After launch, KV-cache storage and reuse are controlled by inserting delimiter tokens into requests sent to the server.
For example, an input of the form \texttt{<delim>\{doc\}<delim>} computes and stores the KV cache for \texttt{\{doc\}}.\footnote{\texttt{<|reserved\_200016|><|reserved\_200017|>} is used as the delimiter token in our current implementation.}
A later request \texttt{<delim>\{doc\}<delim>\{q\}} reuses that cache and performs prefill only for \texttt{\{q\}} onward.
Similarly, in \texttt{\{system\}<delim>\{doc\}<delim>\{q\}}, the server first prefills \texttt{\{system\}}, then reuses the cached KV states for \texttt{\{doc\}}, and finally continues computation for \texttt{\{q\}}.
This allows cache reuse even when the reusable segment is preceded by other prefix content, unlike conventional exact-prefix caching pipelines.

\subsection{Proxy Server}
The proxy server was originally designed to support the evaluation of serving metrics such as TTFT and throughput, but it also serves as a prototype of a deployable front-end for practical services.
Acting as an intermediary between users and the vLLM server, it handles request and response exchange and provides functionalities including pseudo-retrieval, prompt building, and KV-cache management to emulate RAG-style scenarios.

\paragraph{Pseudo-Retrieval}
In a standard RAG pipeline, a user question is first used to retrieve relevant documents from an external knowledge source.
Because our focus is not retrieval quality itself but the behavior of KV-cache reuse under retrieval-augmented workloads, the proxy server implements a pseudo-retrieval mechanism instead of a full retrieval system, as described in Appendix~\ref{sec:database}.
Given a question, the pseudo-retrieval component retrieves its associated document(s) from a preconstructed database and passes them to the prompt-building component within the proxy, enabling controlled and reproducible evaluation under RAG-like request patterns.

\paragraph{Prompt Building}
\label{sec:prompt}
Based on the retrieved document and the current cache state, the proxy constructs the input prompt sent to the vLLM server.
If no KV cache is available for the retrieved document, the proxy first issues a document-only request of the form
\texttt{\{system\}<delim>\{doc\}<delim>} 
where \texttt{\{system\}} is a dedicated system prompt used for KV-cache generation and storage, to generate and store the corresponding KV states.
If the document cache is already available, the proxy sends a request of the form
\texttt{\{prefix\}\{perturbation\}<delim>\{doc\}\\<delim>\{q\}},
where \texttt{\{prefix\}} denotes user-specified text preceding the document, such as a system prompt, and \texttt{\{perturbation\}} is an optional string used to emulate cases in which conventional prefix caching cannot be applied.
In our experiments, it consists of dynamic content such as timestamps.


When the document cache is absent, the proxy could generate it while processing the first full user request. We deliberately avoid this optimization because preceding user-specific segments may influence the cached KV representations, creating a risk of cross-segment information leakage. Instead, the proxy generates reusable document caches through a dedicated request whose prefix contains no user-specific content. Appendix~\ref{sec:cross_segment_leakage} provides an empirical example of this behavior.

\begin{table*}[t]
  \centering
  \begin{tabular}{l|cccc|ccc}
    \hline
    Task (Metrics) & \multicolumn{4}{c}{QA (Token-level F1 score)} & \multicolumn{3}{|c}{Summary (ROUGE-L)}\\
    \hline
    \# Document & Single & \multicolumn{3}{|c}{Multi} & \multicolumn{2}{|c}{Single} & \multicolumn{1}{|c}{Multi}\\
    \hline
    Dataset & QASPER & HQA  & 2WMQA & MuSiQue & QMSum & GovRepo & M-News\\
    \hline\hline
    w/o context&0.04 &0.11 &0.12 &0.11 &0.07 &0.03 &0.04 \\
    Recompute&\textbf{0.42} &\textbf{0.71} &\textbf{0.80} &\textbf{0.61} &\textbf{0.20} &\textbf{0.26} &\textbf{0.22} \\
    Repurp&\textbf{0.42} &0.60 &0.50 &0.40 &0.19 &\textbf{0.26} &0.17 \\
    Repurp (Opus 3)&\textbf{0.42} &0.59 &0.50 &0.41 &0.19 &0.21 &0.17 \\
    Repurp (Haiku 3.5)&\textbf{0.42} &0.56 &0.48 &0.41 &0.19 &0.19 &0.18 \\
    \hline
  \end{tabular}
  \caption{\label{tab:acc}
    Performance comparison of KV-Repurpose and KV recomputation on LongBench datasets.
  }
\end{table*}


\paragraph{KV-Cache Management}
The proxy server also manages the capacity of stored KV caches.
A storage limit can be specified, and when the occupied capacity exceeds this limit, cached entries are evicted using a simple least recently used (LRU) policy.

\paragraph{Usage}
To use the proxy server, the pseudo-retrieval database is first initialized by registering question--context pairs.
The proxy is then launched with a specified cache capacity.
Once running, users only need to submit questions to the proxy, which automatically performs pseudo-retrieval, prompt construction, KV-cache handling, and communication with the vLLM server.

\section{Related work}
\subsection{KV-Cache Management for LLM Inference}
LMCache and several related systems are primarily designed for prefix caching that assumes an exact match between the cached and requested prefixes~\cite{lmcache,redknot,minipic}. 
More specifically, LMCache divides an input sequence into chunks and searches for cached KV states from the first chunk onward. 
Once a cache miss is encountered, subsequent chunks are no longer checked for cache reuse, and the remaining KV states are generated through a new prefill computation.\footref{issue} 
While this behavior is well suited to prefix caching, it is less appropriate for our setting, where later chunks may still be reusable even if an earlier chunk does not yield a cache hit.

Another practical difference lies in system complexity. 
Most existing systems incorporate highly optimized implementations, but this can make the system harder to analyze and extend for research purposes.
In contrast, our extensions are implemented in fewer than 900 lines of Python code, keeping the implementation simple and facilitating reproducibility.

\subsection{Position-Independent Caching}
Our work is also related to recent studies on position-independent caching (PIC), including methods such as CacheBlend~\citep{cacheblend} and EPIC~\citep{epic}, as well as fine-tuning-based approaches~\citep{blockattention,kvlink,turborag,c2kv} for improving robustness to positional mismatch. 
Prior PIC studies mainly focus on preserving output quality under positional mismatch, whereas our work emphasizes practical serving benefits by increasing opportunities for KV reuse when exact prefix matching fails.
The two directions are therefore complementary rather than mutually exclusive, and they may also be combined within a single system to leverage the advantages of both, although such a setting is not considered in this paper.

Another key difference lies in the target model and the intended usage scenario.
Prior PIC studies have mainly evaluated relatively small models and workloads composed of many short document segments. 
In contrast, we focus on a model with more than 100 billion parameters; none of the prior studies discussed above evaluated models at this scale. 
We also consider workloads involving one or a few substantially longer documents, such as full research papers or entire web pages.
This design choice is motivated by prior findings that such a retrieval setting can improve RAG performance~\citep{stone}. 


\section{Evaluation}
\label{sec:eval}
\subsection{Experimental Setup}
\label{sec:setup}
We implement a KV-Repurpose-based testbed as an extension to vLLM v0.20.0.
For all experiments, we use GPT-OSS-120B~\citep{gptoss120b}, served with vLLM on a single NVIDIA H200 GPU, with the KV cache stored in NVMe SSD storage.
For comparison with prefix caching, we evaluate a variant of our system that uses LMCache as its KV connector.

For accuracy evaluation, we use LongBench Pro benchmark~\citep{longbenchpro} and seven datasets from LongBench~\citep{longbench}:
QASPER~\citep{qasper},
QMSum~\citep{qmsum},
GovReport~\citep{govreport},
HotpotQA~\citep{hotpotqa},
2WikiMultihopQA~\citep{2wiki},
MuSiQue~\citep{musique},
and Multi-News~\citep{multinews}.
For system-level evaluation, we construct workloads using English documents ranging from 8K to 128K tokens from the T5.1 Full-Sentence Citation Alignment task in LongBench Pro.

\subsection{Task Performance}
\label{sec:task_p}
For KV-Repurpose, we issue requests to the vLLM server using several system prompts, all of which differ from the prompt used to generate the documents' KV caches.
Specifically, the default system prompt contains 80 tokens, whereas the Claude 3 Opus and Claude 3.5 Haiku prompts from~\citet{claudeprompts} contain 496 and 2,016 tokens, respectively.
We use LMCache as the prefix-caching baseline, which cannot reuse the context KV due to prefix mismatches and therefore recomputes the exact KV states, serving as the quality-preserving reference condition. 
In the without-context condition, the model receives the task instruction and question but not the retrieved document context. 

Following the standard LongBench evaluation protocol, we use the default metrics: token-level F1 for QA tasks and ROUGE-L for summarization tasks.
As shown in Table~\ref{tab:acc}, naive KV-Repurpose provides a useful baseline and is already effective for a meaningful subset of long-context workloads, while revealing substantial limitations in cross-document reasoning settings.
The scores differ substantially from those of the without-context variant, suggesting that the model relies on the provided context rather than answering solely from its internal knowledge.
As an additional diagnostic, we restrict the inputs of the multi-document benchmarks to relevant documents only. 
This filtering partially recovers performance but does not eliminate the degradation, leaving our overall conclusion unchanged. 
We report the detailed results in Appendix~\ref{sec:relevant_document_analysis}.
Table~\ref{tab:acc_pro} reports the average performance on LongBench Pro using its standard evaluation metrics. 
KV-Repurpose and KV recomputation achieve comparable average performance across the 25 tasks. 
Detailed task-level results are provided in Appendix~\ref{app:acc_pro}.
For each question, we cache the entire context as a single KV entry and evaluate KV-Repurpose, following the LongBench single-document setting.
All scores reported in these experiments are averaged over three runs.

\begin{table}[t]
  \centering
  \begin{tabular}{l|c}
    \hline
    Dataset & LongBench Pro (Ave.)\\
    \hline\hline
    w/o context&0.11\\
    Recompute&\textbf{0.45}\\
    Repurp&0.43\\
    Repurp (Opus 3)&\textbf{0.45}\\
    Repurp (Haiku 3.5)&0.44\\
    \hline
  \end{tabular}
  \caption{\label{tab:acc_pro}
    Comparison of the average performance of KV-Repurpose and KV recomputation across the 25 LongBench Pro tasks.
  }
\end{table}

\begin{figure}[t]
  \centering
  \includegraphics[width=0.9\columnwidth]{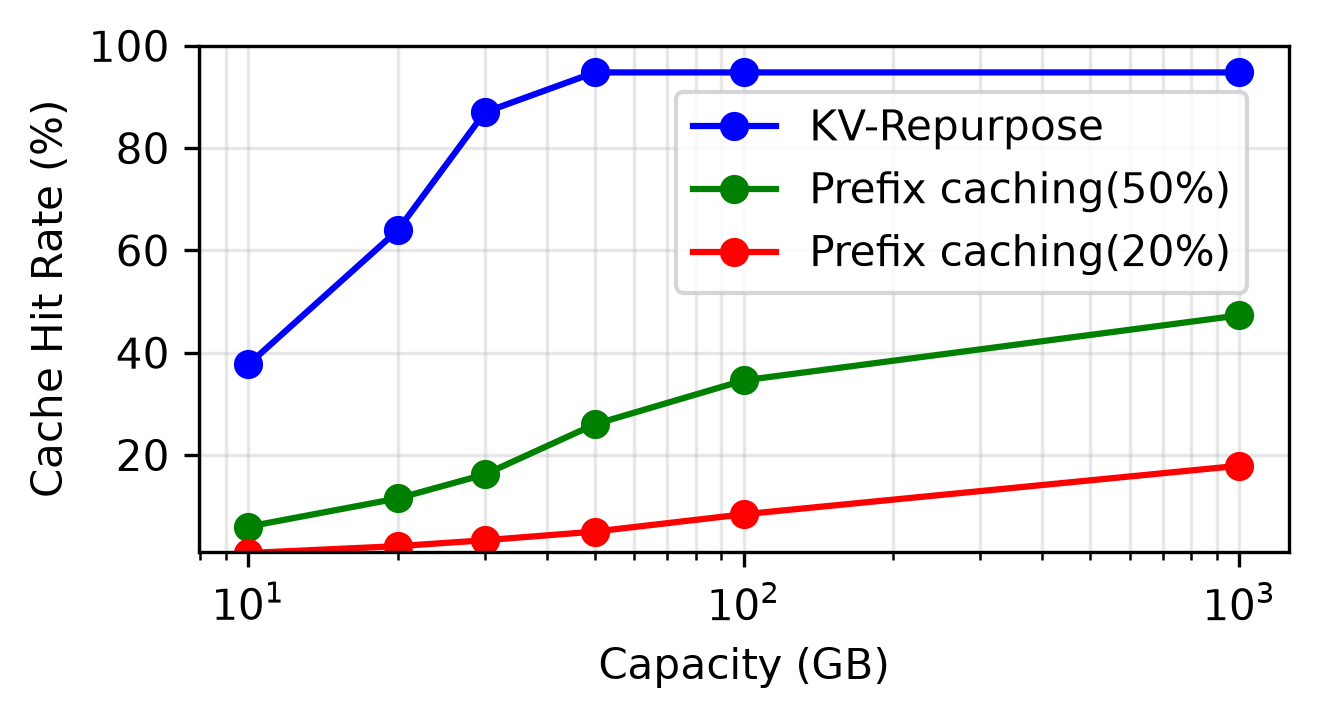}
  \caption{Token-level cache hit rate vs storage capacity.}
  \label{fig:cache_hit_rate}
\end{figure}

\begin{figure}[t]
  \centering
  \includegraphics[width=0.9\columnwidth]{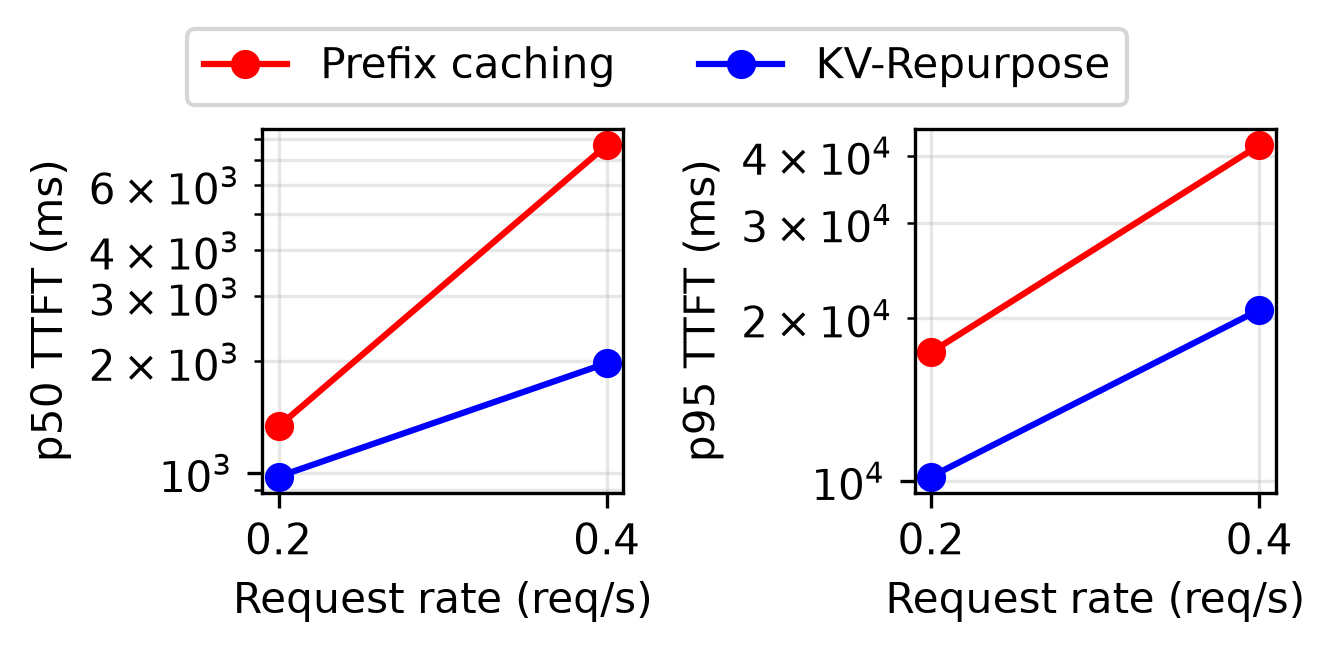}
  \caption{p50/p95 TTFT under a Zipfian workload with Poisson request arrivals and a 30 GB cache capacity.}
  \label{fig:ttft_zipf}
\end{figure}

\begin{figure*}[t]
  \centering
  \includegraphics[width=0.85\linewidth]{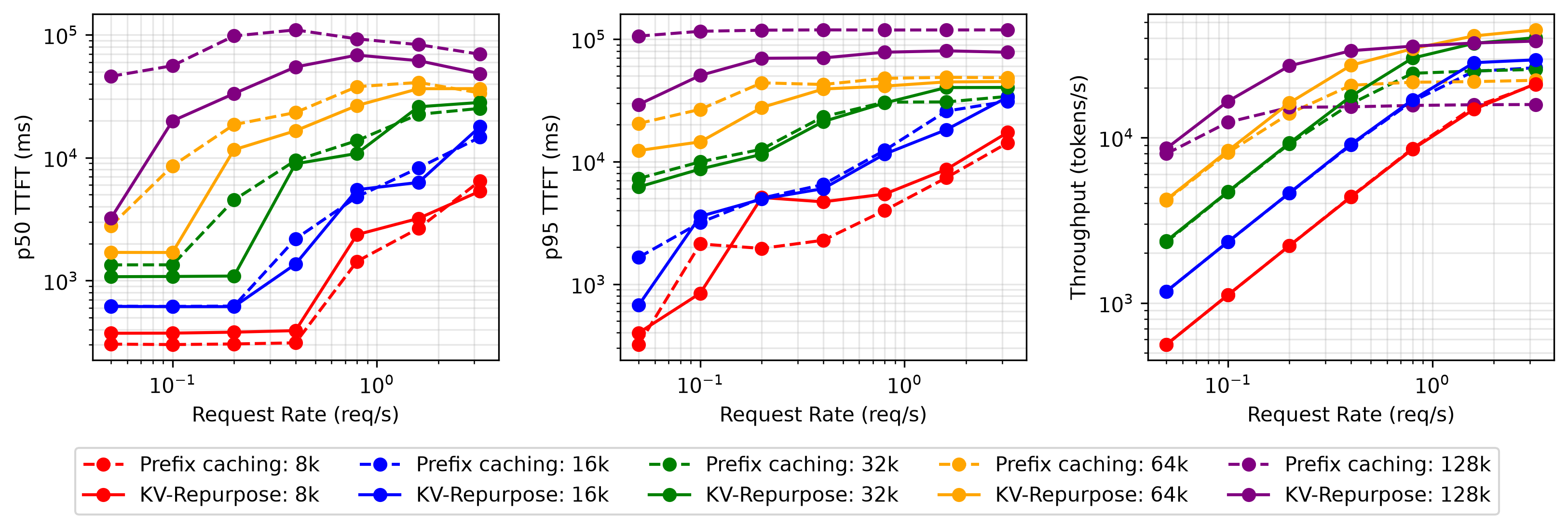}
  \caption {p50 TTFT, p95 TTFT, and throughput across request rates and context lengths.}
  \label{fig:main}
\end{figure*}

\subsection{Evaluation under a Zipfian Workload}
\label{sec:speed}

We evaluate the storage efficiency of KV-Repurpose under a Zipfian distribution~\citep{zipf, web}, in which a small number of documents are accessed frequently.
Conventional prefix caching may store redundant KV states when the same document follows different prefixes, such as different system prompts, conversation histories, or dynamic metadata.
For example, if the same 64K-token document is requested with 100 distinct prefixes, prefix caching may store up to 100 prefix-dependent copies of its KV states, whereas KV-Repurpose stores one document cache shared across the requests.
Although such skewed reuse does not characterize every workload, it represents a plausible setting in which eliminating prefix-dependent duplication can provide substantial capacity savings.
We use it to illustrate a practically relevant case where KV-Repurpose is particularly effective.
 
We construct the workload by sampling 2,500 requests from LongBench Pro T5.1 according to a Zipfian distribution over documents.
In two configurations, 20\% or 50\% of requests use a fixed datetime prefix, while the remainder use request-specific timestamps.
Thus, exact prefix reuse is available only for requests sharing the fixed prefix, whereas KV-Repurpose can reuse document KVs across all timestamp variants.
As shown in Figure~\ref{fig:cache_hit_rate}, KV-Repurpose consistently achieves substantially higher token-level cache hit rates than prefix caching.
At the same cache capacity of 30~GB, KV-Repurpose achieves a 63.9\% hit rate, compared with 3.3\% and 16.2\% for prefix caching in the 20\% and 50\% configurations, respectively. Notably, prefix caching reaches only 17.9\% and 47.3\% even with a 1-TB cache, highlighting the superior capacity efficiency of KV-Repurpose.
Figure~\ref{fig:ttft_zipf} reports the TTFT results with a 30 GB cache under the same workload, where request arrivals follow a Poisson process. Owing to the substantial difference in cache hit rate, KV-Repurpose achieves lower p50 and p95 TTFT than prefix caching.
This advantage is expected to remain significant even when extrapolated to much larger-scale datasets. (Appendix~\ref{sec:real})

\subsection{Context-Length Scaling Microbenchmark}
\label{sec:ttft}
To isolate the effect of context length from document popularity and cache-eviction behavior, we conduct a controlled microbenchmark using a fixed context at each evaluated length. Each context is requested 100 times, allowing us to compare recomputation, prefix caching, and KV-Repurpose under a stable warm-cache condition.
In this experiment, the request time is prepended to the prompt template described in Section~\ref{sec:prompt} as a perturbation prefix, and the request arrival process is assumed to follow a Poisson distribution.
We vary the request rate in requests per second (RPS), and evaluate latency and throughput using TTFT and tokens per second, respectively. 
Both metrics are computed over the complete request stream, including requests that trigger initial cache generation.
We provide the detailed processing flow in Appendix~\ref{sec:processing_flow}.
 
Figure~\ref{fig:main} shows TTFT and throughput as functions of context length and request rate for prefix caching and KV-Repurpose.
When the context length exceeds roughly 16K tokens, KV-Repurpose achieves lower TTFT owing to its increased opportunities for KV-cache reuse.
In the throughput experiments, the prefix-caching approach saturates at high request rates, whereas the proposed system sustains higher throughput across most evaluated high-load settings.
These results highlight the importance of KV-cache reuse in reducing GPU load: when cached KVs cannot be reused, repeated prefill computation consumes GPU resources, preventing the system from keeping up with high request rates.
By expanding reuse opportunities beyond prefix caching, KV-Repurpose avoids more prefill computation and frees additional GPU resources, allowing the system to sustain higher request rates.

\section{Conclusion}
In this paper, we introduced KV-Repurpose and investigated its potential for reducing the prefill cost of LLM inference.
We implement a testbed that supports flexible KV-cache reuse in an LLM inference engine and evaluate it under controlled RAG-style serving scenarios.
By reducing redundant KV-cache storage, our system improves storage efficiency, and by increasing cache hit rates, it improves response performance while reducing GPU computational load. 
The storage-efficiency advantage of KV-Repurpose may become increasingly important as KV-cache repositories grow, although evaluating such large-scale deployments is left for future work.
Overall, these gains substantially broaden the benefits of KV-cache optimization for LLM inference by enabling the retention of more useful and diverse KV entries, and may even open up new application opportunities.
\section*{Limitations}
This study aims to provide a simple testbed for investigating the effectiveness of KV-Repurpose rather than exhaustively exploring all possible techniques.
Our implementation applies only a simple positional correction and neither combines KV-Repurpose with nor directly compares it with the more sophisticated techniques proposed in prior PIC studies.
Moreover, our experiments use only a single LLM, so the results may not generalize to other model architectures, sizes, or training settings.
Evaluating KV-Repurpose across a broader range of models and workloads and comparing or combining it with existing PIC techniques remain important directions for future work.

\bibliography{custom}
\newpage
\appendix

\section{Appendix}
\label{sec:appendix}




\subsection{Database}
\label{sec:database}

For each question, we create a database entry that contains the gold context provided in LongBench Pro.
The context associated with each question is then regarded as the retrieved and reranked result and fed into the prompt processing tool.
Construction of the database is shown as Algorithm \ref{alg:build_pseudo_db}.

\begin{algorithm}[h]
\caption{Construction of the pseudo-retrieval database}
\label{alg:build_pseudo_db}
\begin{algorithmic}[1]
\Require Question set $Q$, gold context mapping $G$, context pool $P$
\Ensure Pseudo-retrieval database $D$
\State $D \leftarrow \emptyset$
\ForAll{$q \in Q$}
    \State $D[q] \leftarrow G[q]$
\EndFor
\State \textbf{return} $D$
\end{algorithmic}
\end{algorithm}

\subsection{Empirical Observation of Cross-Segment Information Leakage}
\label{sec:cross_segment_leakage}

Although a detailed investigation of KV-cache security is outside the scope of this paper, we briefly report an empirical observation that influenced the cache-generation design adopted in KV-Repurpose.

During system development, we observed that information appearing before a cached context segment could occasionally be reproduced after the corresponding KV cache was reused.
To evaluate this behavior, we prepared a set of 100 popular family names consisting of 50 common American surnames and 50 common Japanese surnames.

For each name, we generated a KV cache using the following prompt:

\begin{quote}
My name is \{name\}. \{delimiter\}\{context\}\\
Here please remind my name. Then please say my name after [Answer].\{delimiter\}
\end{quote}

The reusable KV cache was created only from the \texttt{\{context\}} segment. Consequently, the family name itself was not included in the cached segment and appeared only in the preceding prompt segment during KV-cache generation.

The generated KV cache was then reused to answer a question about the context using the following prompt:

\begin{quote}
\{delimiter\}\{context\}\{delimiter\}\\
\{question\}\\
Here please remember my name. Then please say my name after [Answer]
\end{quote}

Each name was evaluated three times. A name was counted as leaked if it appeared correctly in the model output in at least one of the three trials. Among the 100 names, leakage was observed for 35 names (35.0\%).

As a control experiment, we generated KV caches without the name-containing prefix:

\begin{quote}
\{delimiter\}\{context\}\\
Here please remind my name. Then please say my name after [Answer].\{delimiter\}
\end{quote}

In this setting, no name leakage was observed.

These results suggest that information contained in prompt segments preceding a cached context may be implicitly reflected in the resulting KV representations. As a consequence, if reusable KV caches are generated from prompts containing user-specific or sensitive information, such information may later be reproduced when the cache is reused, even though the information is not explicitly present in the cached segment itself.

This observation motivated the cache-generation strategy adopted in KV-Repurpose. To avoid potential security and privacy risks, reusable document KV caches are generated from the document context alone as a dedicated preprocessing step. User-specific prefixes and other preceding prompt segments are excluded from cache generation, ensuring that reusable caches do not inadvertently inherit information from unrelated requests.

\subsection{End-to-End Workflow for Subsection~\ref{sec:speed} and~\ref{sec:ttft}}
\label{sec:processing_flow}



\begin{algorithm}[h]
\caption{Proxy-server workflow for \ref{sec:ttft}}
\label{alg:speed_flow}
\begin{algorithmic}[1]
\Require System prompt \texttt{sys}, perturbation prefix \texttt{P}, context \texttt{C}, question \texttt{Q}, cache store $M$, cache capacity $\Gamma$, mode $m \in \{\textsc{prefix-cache}, \textsc{repurpose}\}$
\State Construct full prompt $X_{\text{full}} \leftarrow [\texttt{sys}; \texttt{P}; \texttt{C}; \texttt{Q}]$
\If{$m = \textsc{prefix-cache}$}
    \State Send $X_{\text{full}}$ to vLLM w/ LMCache
\ElsIf{$m = \textsc{repurpose}$}
    \If{$M$ contains KV cache for context $C$}
        \State Send $X_{\text{full}}$ to our vLLM
    \Else
        \State Construct context-only prompt 
        \Statex \hspace{\algorithmicindent}\hspace{\algorithmicindent} $X_{\text{context}} \leftarrow [\texttt{sys}; \texttt{C}]$
        \State Send $X_{\text{context}}$ to our vLLM
        \State Send $X_{\text{full}}$ to our vLLM
    \EndIf
\EndIf
\If{$\mathrm{Size}(M) > \Gamma$}
    \State Evict least-recently-used entries from $M$
    \Statex \hspace{\algorithmicindent} until $\mathrm{Size}(M) \le \Gamma$
\EndIf
\end{algorithmic}
\end{algorithm}

The workload emulator sends requests to the proxy server at times sampled from a Poisson process.
The request rate is measured based on these request arrival times.
After receiving a request, the proxy server performs pseudo-retrieval to obtain a context-question pair.
The request is then sent to the vLLM server following the procedure described in Algorithm~\ref{alg:speed_flow}, and outputs for each mode are generated.
In addition, the proxy server updates the metadata stored in the KV management database.
When necessary, it evicts KV cache from storage according to the least recently used (LRU) policy.
Here, TTFT denotes the time until the first reasoning token is generated.
For each context length in Figure~\ref{fig:main}, we use one fixed context and process the corresponding request 100 times.


\subsection{Illustrative Capacity Extrapolation}
\label{sec:real}

\begin{figure}[t]
  \centering
  \includegraphics[width=0.85\linewidth]{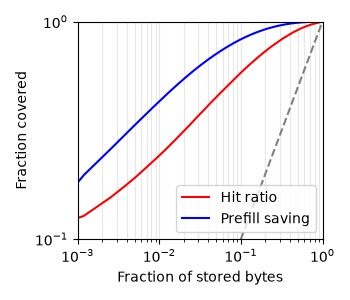}
  \caption {Storage vs hit ratio and prefill saving.}
  \label{fig:wiki}
\end{figure}

Figure~\ref{fig:wiki} shows the relationship between the fraction of KV-cache capacity and the cache hit rate for English Wikipedia pages that received at least one page view during a single day~\citep{wikimedia_pageview_data}.
The total KV-cache size required to store all such pages is normalized to 1, and pages are sorted in descending order of access frequency.
The x-axis represents the fraction of the total KV-cache capacity allocated to the most frequently accessed pages, while the y-axis shows the resulting cache hit rate if those pages are cached.
A cache capacity corresponding to only 0.2\% of the total achieves a cache hit rate of 14.55\%, while 1\% of the total capacity achieves 24.2\%.
KV-Repurpose can directly benefit from this cache hit rate.
Assuming that the KV-cache size is proportional to the Wikipedia page size, with a proportionality constant of 15 KB of KV cache per byte of Wikipedia content, a cache capacity of 0.2\% corresponds to approximately 1.7 TB.
The Wikipedia page-view dataset also exhibits a positive correlation between page size and access frequency.
Therefore, if the amount of prefill computation saved is assumed to be proportional to the product of the cache hit rate and the cached KV capacity, the estimated prefill savings reach 23.74\% with 0.2\% cache capacity and 42.94\% with 1\% capacity.

In contrast, prefix caching can reuse a cached KV only when the prompt prefix is identical.
Even if the same Wikipedia page is referenced, the cached KV cannot be reused when the prompt prefix differs.
Consequently, in practice, the KV cache for a Wikipedia page is reusable only when the same user continues a multi-turn conversation that repeatedly references that page.
As a simplifying assumption, we assume that at most 20\% of page views correspond to the same user revisiting the same page, which we consider a generous upper bound given the usage characteristics of Wikipedia.
Assuming that this revisit probability is independent of page size, both the cache hit rate and the prefill savings are therefore upper-bounded by 20\%.

Furthermore, even under assumptions that are highly favorable to prefix caching---namely, (1) the system can perfectly predict which requests will reuse a Wikipedia page, and (2) every conversation consists of 10 turns---the required KV-cache capacity would still be approximately 2.8 PB, calculated as the cumulative page size over all accesses $\times$ 15 KB $\times$ 20\% $\times$ 1/10.

\begin{figure*}[t]
  \includegraphics[width=\linewidth]{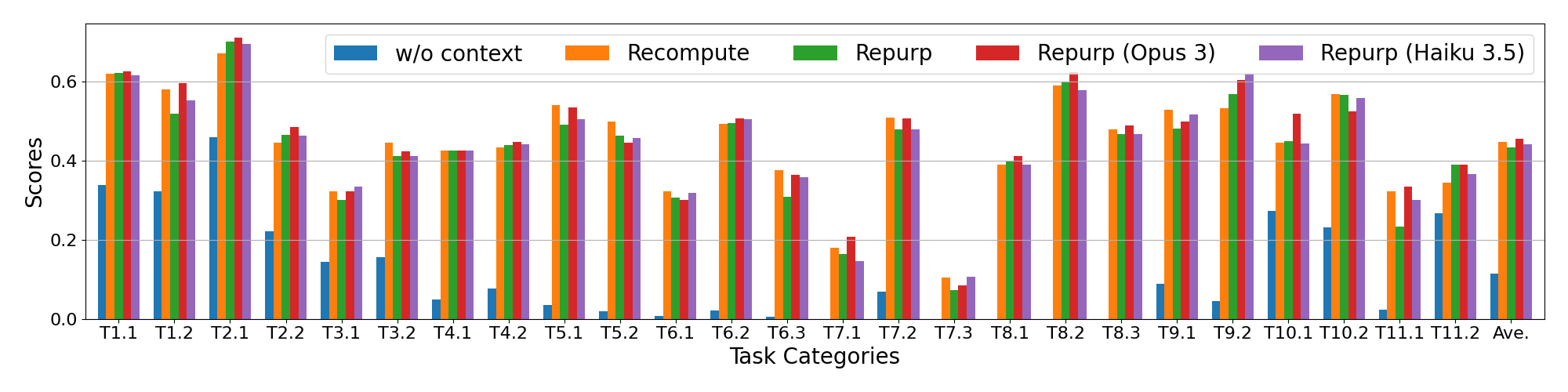}
  \caption {Performance comparison of KV-Repurpose and KV recomputation on LongBench Pro datasets.}
  \label{fig:acc_pro}
\end{figure*}

The numerical assumptions used in this extrapolation, including 15 KB of KV-cache storage per byte of document content and a 20\% reuse probability, are intended only to make the scale of the illustrative scenario easier to interpret. We do not claim that these values precisely characterize a particular production workload or deployment. Nevertheless, the qualitative conclusion does not depend on their exact values. Prefix caching may retain multiple prefix-dependent KV-cache entries for the same document when it is requested under different system prompts, conversation histories, or other preceding contexts. In contrast, KV-Repurpose allows these requests to share a single document cache. Consequently, for the same KV-cache capacity, KV-Repurpose can avoid substantially more document-prefill recomputation in workloads where documents are repeatedly accessed under diverse prefixes.


\begin{table}[t]
  \centering
  \begin{tabular}{l|ccc}
    \hline
    Dataset & QA1  & QA2 & QA3\\
    \hline\hline
    Recompute&0.71 &0.78 &0.60 \\
    Repurp&0.67 &0.57 &0.47 \\
    Repurp (Opus 3)&0.68 &0.59 &0.47 \\
    Repurp (Haiku 3.5)&0.68 &0.56 &0.49 \\
    \hline
  \end{tabular}
  \caption{\label{tab:supp}
    Performance comparison of KV-Repurpose and KV recomputation on HotpotQA (QA1), 2WikiMultihopQA (QA2), and MuSiQue (QA3), using only supporting passages as input documents.
  }
\end{table}

\subsection{Evaluation with Selected Supporting Documents}
\label{sec:relevant_document_analysis}

To further examine the performance degradation on the multi-document QA tasks, we conduct an additional diagnostic evaluation using a reduced set of input documents.
For each instance, we provide GPT-OSS-120B with the question, reference answer, and candidate documents, and ask it to select the documents relevant to answering the question.
This procedure typically retains between two and several documents per instance.
Because the selection uses the reference answer, this experiment should be viewed as an oracle-assisted diagnostic rather than a standard evaluation setting.

Table~\ref{tab:supp} reports the results.
Compared with the original multi-document setting in Table~\ref{tab:acc}, restricting the input to the selected documents partially reduces the performance degradation of KV-Repurpose, although a performance gap remains.
Because this intervention simultaneously reduces the number of concatenated documents and removes potentially irrelevant content, we do not attribute the recovery to either factor individually.
Overall, the result does not change our main conclusion that multi-document settings remain a limitation of naive KV-Repurpose.

\subsection{Detailed Results on LongBench Pro}
\label{app:acc_pro}

Figure~\ref{fig:acc_pro} reports the task-level results for all 25 tasks in LongBench Pro. Following the standard evaluation protocol, we use the metric specified for each task. 
KV-Repurpose and KV recomputation exhibit minor task-level variations while achieving comparable average performance. 
In addition, the without-context variant generally yields substantially lower scores, providing further evidence that the model makes effective use of the provided context.
\end{document}